%% file: root.tex
\documentclass[letterpaper, 10 pt, conference]{ieeeconf}
\IEEEoverridecommandlockouts
\input{macro}

\title{\LARGE \bf
\resizebox{\linewidth}{!}{
\method: A Hand-Held Skill Interface for Dexterous Hand Data Collection
}
}

\author{Amber Xie$^{1,2}$, Jiawei Fu$^{1,3}$, Xiaofeng Guo$^{1}$, Carmelo Sferrazza$^{\dagger1}$,
Guanya Shi$^{\dagger1}$, and Haozhi Qi$^{\dagger1}$\\
$^{1}$Amazon FAR (Frontier AI \& Robotics), $^{2}$Stanford University, $^{3}$UC San Diego%
\thanks{$^{\dagger}$Amazon FAR team co-lead.}%
}

\begin{document}

\let\originaltwocolumn\twocolumn
\renewcommand\twocolumn[1][]{%
    \originaltwocolumn[{#1}{%
        \input{sections/0_teaser}
    }]%
}
\maketitle
\let\twocolumn\originaltwocolumn
\thispagestyle{empty}
\pagestyle{empty}

\begin{abstract}

Collecting high-quality demonstrations remains a central bottleneck for learning
dexterous manipulation. Teleoperation records executable robot actions, but it
requires a complete robot setup and
places the burden of coordinating many finger joints on the operator. Human
demonstrations are easier to collect, but their motions must be estimated and
retargeted to the robot, and they do not directly provide robot actions. In this
paper, we introduce \method, a portable data-collection interface for high-DoF
robot hands. It uses the same hand hardware for collection and deployment,
avoiding a hand embodiment gap, and records both commanded joint targets and
measured joint states. Our key
observation is that a small set of skills can support a wide range of dexterous
tasks. During collection, the operator moves the hand-held interface to specify
the robot's wrist motion and selects or adjusts skills that coordinate the finger
motion. This removes the need to control
each joint independently while retaining behaviors such as precise grasping and
in-hand manipulation. We collect over 2,500 demonstrations across several real-world
tasks in laboratory and in-the-wild settings without a robot arm. We then use behavior
cloning to train visuomotor policies directly on the recorded wrist motions and
hand commands, without skill labels. 
We evaluate on four tasks: pick-and-place with diverse objects, screwing with
in-hand tool use, drawer stowing that requires simultaneous grasping and pulling,
and bimanual barcode scanning. Using recorded robot commands instead of measured
joint states improves average final-task success from 11\% to 64\% across the
three tasks evaluated by success rate. Co-training with in-the-wild data improves
average pick-and-place success from 36\% to 78\% across the three evaluation
conditions. Videos are available at https://amberxie88.github.io/dexport/.
\end{abstract}

\input{sections/1_intro}
\input{sections/2_related}
\input{sections/3_method}
\input{sections/4_experiments}
\input{sections/5_conclusion}

\bibliographystyle{IEEEtran}
\bibliography{references}

\input{sections/6_appendix}

\end{document}

%% file: macro.tex
\usepackage{graphicx}
\usepackage{amsmath}
\usepackage{amssymb}
\usepackage{xcolor}
\usepackage{booktabs}
\usepackage{xspace}
\usepackage{capt-of}
\usepackage{wrapfig}
\usepackage{pifont}
\providecommand{\Checkmark}{\ding{51}}
\providecommand{\XSolidBrush}{\ding{55}}

\makeatletter
\let\original@makecaption\@makecaption
\long\def\@makecaption#1#2{%
    \ifx\@captype\@IEEEtablestring
        {\footnotesize\normalfont\raggedright\noindent #1: #2\par}%
        \vskip 6pt
    \else
        \original@makecaption{#1}{#2}%
    \fi
}
\makeatother

\usepackage{cleveref}
\crefname{section}{Sec.}{Secs.}
\Crefname{section}{Sec.}{Secs.}
\crefname{subsection}{Sec.}{Secs.}
\Crefname{subsection}{Sec.}{Secs.}
\crefname{subsubsection}{Sec.}{Secs.}
\Crefname{subsubsection}{Sec.}{Secs.}
\crefname{table}{Table}{Tables}
\Crefname{table}{Table}{Tables}
\crefname{figure}{Fig.}{Figs.}
\Crefname{figure}{Fig.}{Figs.}
\crefname{appendix}{App.}{Apps.}
\Crefname{appendix}{App.}{Apps.}
\crefname{subappendix}{App.}{Apps.}
\Crefname{subappendix}{App.}{Apps.}
\crefformat{equation}{(#2#1#3)}
\Crefformat{equation}{(#2#1#3)}
\crefrangeformat{equation}{(#3#1#4)--(#5#2#6)}
\crefmultiformat{equation}{(#2#1#3)}{ and (#2#1#3)}{, (#2#1#3)}{, and (#2#1#3)}

\definecolor{darkblue}{HTML}{273A5A}

\definecolor{orange}{HTML}{FF7F00}

\definecolor{ccpurple}{HTML}{7B2D8E}

\newcommand{\noindentparagraph}[1]{%
    \vspace{0.2em}\noindent\textbf{#1}%
}

\newcommand{\method}{\textsc{DexPort}\xspace}

%% file: sections/0_teaser.tex
\begingroup
\centering
\includegraphics[width=\linewidth]{fig/main.pdf}
\captionof{figure}{\textbf{Portable dexterous data collection with \method.}
(a) The operator moves the hand-held device to control wrist motion and uses a
small set of skills to control the fingers. The portable interface uses the
same robot hand used at deployment and records commanded joint targets and measured
joint states. (b) Visuomotor policies trained on these demonstrations predict
wrist and hand commands directly, supporting varied grasps, bimanual
coordination, in-hand manipulation, and opening a drawer while holding an object.}
\label{fig:main}
\vspace{1mm}
\par
\endgroup

%% file: sections/1_intro.tex
\section{INTRODUCTION}
\label{sec:intro}

Multi-fingered hands allow robots to use different grasps, manipulate tools, and
reorient objects in hand. Learning these behaviors from demonstrations requires
data that capture detailed finger motion and contact-rich interactions.
Teleoperation records executable robot actions, but it requires a complete robot
setup. Collecting in-hand manipulation demonstrations through teleoperation is
particularly challenging because the operator must coordinate many finger
joints~\cite{dexpilot,anyteleop,telekinesis,anydexrt,ace}. On the other hand, human hand
demonstrations are easier to collect, but their motions must be estimated and
retargeted to the robot, and they do not directly provide robot
actions~\cite{dexcap,dexwild,aina}. We therefore need a method that scales beyond
a fixed robot setup while preserving executable hand actions.

To address this issue for parallel-jaw grippers, the Universal Manipulation Interface (UMI)~\cite{umi} uses a portable, hand-held
interface to collect demonstrations naturally and at scale. However, extending this paradigm to multi-fingered hands has not been
trivial. Early efforts, including DexUMI~\cite{dexumi} and DEXOP~\cite{dexop},
can collect dexterous and contact-rich behaviors using wearable hand exoskeletons.
However, designing a high-DoF exoskeleton involves a difficult trade-off between
dexterity and wearability: supporting more independent finger motions increases
both mechanical complexity and the operator's control burden. In addition, these systems
record exoskeleton configurations rather than the commands and resulting states
of the deployed robot hand. RealDexUMI~\cite{realdexumi} preserves this
information by collecting executable commands with the physical robot hand, but
demonstrates the approach on a hand with only six active DoFs. It remains unclear
how to preserve these benefits for a hand with many independently controlled
joints without making the interface cumbersome.

Inspired by these works, we address three requirements for a portable interface
for dexterous hands. \Cref{tab:interfaces} compares existing systems against these
requirements.
First, the interface should use the same end effector as the deployed robot
(\textbf{Same Emb.}), so that hand motions need no retargeting and no embodiment
gap is introduced. Second, it should record the commands sent to the hand
(\textbf{Action Data}) rather than relying only on actions inferred from future
measured states. This distinction matters during contact, when the commanded
target and measured joint state may differ. Third, it should expose enough
controllable degrees of freedom (\textbf{DoF}) to support tasks that require
multi-fingered dexterity; otherwise, a simpler gripper would be sufficient.
Reusing the deployment hand satisfies the first two requirements, but creating
an intuitive and comfortable interface for high-DoF control is challenging,
especially when the operator must control each joint directly.

\begin{table}[t]
    \vspace{6pt}
    \centering
    \small
    \setlength{\tabcolsep}{3.5pt}
    \begin{tabular}{lcccc}
        \toprule
        System & DoF & Weight & Same Emb. & Action Data \\
        \midrule
        UMI~\cite{umi}               & 1      & 780\,g     & \Checkmark & \XSolidBrush   \\
        DexViTac~\cite{dexvitac}     & 6      & $<$600\,g  & \XSolidBrush   & \XSolidBrush  \\
        DEXOP~\cite{dexop}           & 12/9/7 & $>$1000\,g & \Checkmark & \XSolidBrush  \\
        DexEXO~\cite{dexexo}         & 6      & $>$1000\,g & \Checkmark & \XSolidBrush   \\
        DEX-Mouse~\cite{dexmouse}    & 6      & $>$1000\,g & \Checkmark & \Checkmark \\
        Exo-ViHa~\cite{exoviha}      & 6      & 1100\,g    & \XSolidBrush   & \Checkmark \\
        DexUMI~\cite{dexumi}         & 12/6   & 850\,g     & \Checkmark & \XSolidBrush   \\
        RealDexUMI~\cite{realdexumi} & 6      & 680\,g     & \Checkmark & \Checkmark \\
        \midrule
        \textbf{\method}             & \textbf{20} & 955\,g & \Checkmark & \Checkmark \\
        \bottomrule
    \end{tabular}
    \vspace{0.5\baselineskip}
    \caption{\textbf{Comparison of dexterous demonstration interfaces.}
    Systems are compared by controllable hand degrees of freedom (DoF), device
    weight, use of the deployed robot's hand embodiment (Same Emb.), and recording
    of commanded joint targets (Action Data). \method controls a 20-DoF hand
    and records the commands sent to its joints.
    }
    \label{tab:interfaces}
    \vspace{-20pt}
\end{table}

In this paper, we introduce \method, a portable data-collection interface built
around these requirements. Our key design principle is to collect demonstrations with the same
hand hardware used at deployment while controlling it through a small set of skills.
Because the collection hand shares the deployment hand's hardware and control
interface, each demonstration records native commanded joint targets and
resulting states.
The operator moves the device to specify the robot's wrist motion and selects or
adjusts skills to control the hand. Parameterized skills produce power, pinch, and two-finger
grasps, while a learned closed-loop skill generates the finger gaiting needed
for in-hand rotation. This skill-level interface leaves task decisions to the
operator while the robot hand produces the coordinated finger motion.

We evaluate \method on four tasks that require different forms of dexterity:
selecting grasps for objects of different shapes, opening a drawer while holding
an object in the same hand, rotating a tool in hand, and coordinating two hands
to scan a barcode. We collect demonstrations both in the lab and in
unstructured environments, without a robot arm or external tracking
infrastructure. We then use behavior cloning to train visuomotor policies
directly on the recorded wrist motions and hand commands, without skill labels.
Our policies complete all four tasks and outperform a state-as-action baseline
trained without commanded hand actions on each task. Across the three tasks scored by
success rate, final-task success averages 64\% versus 11\%. Co-training with in-the-wild demonstrations
further improves pick-and-place in every condition, including unseen objects and
scenes with distractors. \method also collects demonstrations at roughly
$1.4\times$ the rate of glove teleoperation.

To summarize, our contributions are as follows:
\begin{itemize}
  \item A portable and ergonomic data-collection interface that uses the same
    hand hardware and control space as the deployed robot. It avoids a hand embodiment
    gap and records native commanded actions and measured states.
  \item A skill-level control scheme that makes a high-DoF hand practical to
    operate during portable data collection. A compact set of parameterized and
    learned skills replaces joint-by-joint control while preserving the ability
    to perform dexterous tasks.
  \item An evaluation across four real-world tasks, including data collection in
    unstructured environments and visuomotor policies trained directly on the
    recorded wrist motions and hand commands, without skill labels.
\end{itemize}

%% file: sections/2_related.tex
\section{RELATED WORK}

\noindentparagraph{Hand-Held Grippers for Robot Learning.}
Hand-held interfaces offer a middle ground between robot teleoperation and
unconstrained human video: they can be carried into diverse environments while
retaining an end-effector-centered observation and action space. Early systems
used instrumented reacher-grabbers or other portable tools and recovered their
motion using structure from motion, visual odometry, or smartphone
odometry~\cite{graspwild,vime,dobbe}. UMI~\cite{umi} makes this approach more
accurate and broadly useful through a hand-held parallel-jaw gripper, a wide
field-of-view camera, visual-inertial pose tracking, and a policy interface that
accounts for deployment latency. Matching the collection and deployment
grippers further reduces the embodiment gap. UMI demonstrates that portable
collection can support precise and long-horizon behaviors, but its gripper has
only one actuated degree of freedom. Extending these benefits to a multi-fingered
hand requires a practical way to control its higher-dimensional action space.

\begin{figure*}[t]
    \vspace{6pt}
    \centering
    \includegraphics[width=\linewidth]{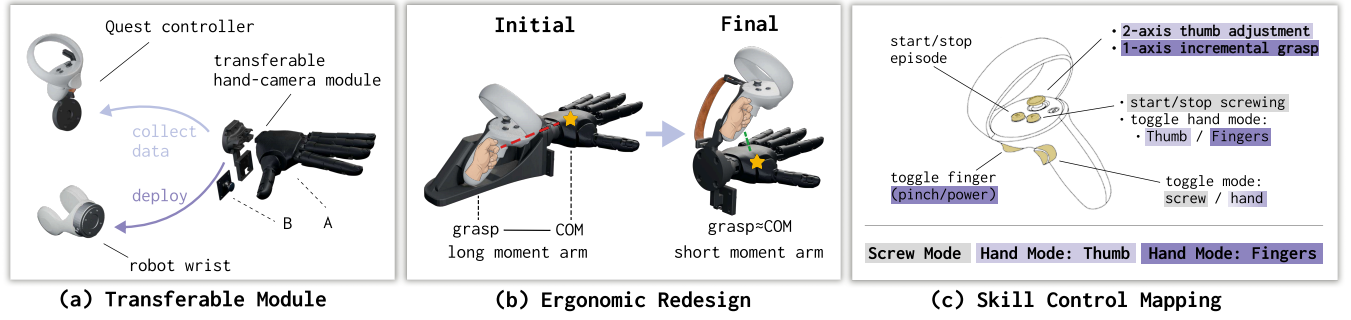}
    \caption{\textbf{Transferable Hardware Interface and Control Mapping.}
    (a) \textit{Transferable Module:}  The Wuji hand (A) and fisheye camera (B) form a transferable hand--camera module that mounts to the Quest 2 controller for data collection and to the robot wrist for deployment. Reusing the same module preserves hand embodiment and camera-to-hand extrinsics.
    (b) \textit{Ergonomic Redesign:} The final hardware interface places the operator's grasp near the assembly's center of mass and adds a support strap, shortening the moment arm and reducing wrist torque.
    (c) \textit{Skill Control Mapping:}  Quest 2 inputs start and stop demonstration episodes, select hand or screw mode, and modulate thumb, grasp, and screwing skills. The joystick adjusts thumb flexion and abduction in Thumb Mode and incrementally tightens or loosens the grasp in Finger Mode.
    }
    \label{fig:primitive-mapping}
    \vspace{-14pt}
\end{figure*}

\noindentparagraph{Learning Dexterous Manipulation.} Dexterous manipulation skills
are commonly learned from simulation, human
demonstrations, or robot teleoperation. Simulation can generate robot actions at
scale and has enabled complex in-hand behaviors, but transferring policies that
rely on contact dynamics and tactile observations to real hardware remains
difficult~\cite{hora,dexscrew,
human2sim2robot}. Human demonstrations collected from videos, hand tracking, or
motion-capture systems provide natural and diverse behaviors~\cite{videodex,
dexmv,telekinesis,dexcap,dexwild,aina}. However, hand motion must be
estimated and retargeted, and the resulting data do not directly contain robot
actions. Teleoperation instead records executable actions by mapping human-hand
motion to a robot hand~\cite{dexpilot,anyteleop,telekinesis,anydexrt,ace,jiang2026concurrent}.
Immersive visual feedback and haptic devices can improve control~\cite{television,
bunnyvision,doglove}, but these systems generally require a complete robot setup
and ask the operator to coordinate many finger joints. Lower-dimensional
interfaces reduce this burden. HATO maps a small set of controller inputs to
coordinated finger motions~\cite{hato}, while DexScrew uses a simulation-trained
rotation policy as a skill during teleoperation~\cite{dexscrew}. These results
motivate skill-level control, but neither system addresses portable collection
with a high-DoF hand.


\noindentparagraph{Portable Data Collection for Dexterous Hands.}
Recent systems enable dexterous data collection without a fixed robot setup.
DexUMI, DEXOP, and DexEXO use wearable exoskeletons or passive hand mechanisms
to constrain and measure the operator's motion~\cite{dexumi,dexop,dexexo}.
These systems efficiently collect demonstrations of contact-rich behaviors and reduce the
human-to-robot kinematic gap. However, these passive collection mechanisms
record hand configurations rather than both the commands and resulting states
of the actuated hand used at deployment. Related systems improve kinematic alignment or tactile sensing
through mechanically isomorphic and modular exoskeleton designs~\cite{mile,
exoviha}. Another line of work uses robot hands that operators carry or wear, reducing
the contact-geometry gap and allowing robot-aligned data
collection~\cite{hirohand,dexmouse}. RealDexUMI~\cite{realdexumi} follows the
UMI principle of using the same end effector for collection and deployment.
It also records robot commands, rather than inferring actions from measured
states, but demonstrates this approach on a hand with six active DoFs.
In contrast, \method targets a 20-DoF hand,
for which joint-by-joint operation would be cumbersome. We use a compact set of
skills to make the hand practical to operate while still recording the raw hand
commands used to train the policy.

%% file: sections/3_method.tex
\section{METHOD}

In ~\cref{sec:hardware}, we introduce the hardware design of our portable data collection interface. Next, in ~\cref{sec:skills} we describe our library of skills, designed to be intuitive and useful for learning dexterous policies. Finally, ~\cref{sec:policy} explains how this data can be used for imitation learning.

\subsection{Hardware Design}
\label{sec:hardware}

\noindentparagraph{Transferable Hand-Camera Module.}
Our portable data-collection interface comprises a transferable hand-camera
module and a collection handle built around a Meta Quest 2 controller, as shown
in \cref{fig:primitive-mapping}(a). The module combines a five-fingered, 20-DoF
Wuji robot hand with a wide-angle fisheye camera through a rigid 3D-printed
flange. During collection, it attaches
to the collection handle; during deployment, the same physical module mounts to
the robot wrist. Reusing the module preserves the hand embodiment and
camera-to-hand extrinsics across the two settings, so recorded hand targets
transfer to the robot without retargeting. This satisfies the \textbf{Same Emb.}
requirement. The mechanical design can be adapted to other
hand-camera pairs by replacing the flange.
We design flanges for both the Wuji 1 and Wuji 2.

\noindentparagraph{Recorded Signals.}
During data collection, the Quest 2 controller provides the wrist's 6-DoF trajectory,
from which we construct end-effector motion targets. The fisheye camera captures
the hand-object interaction and surrounding task context, while the hand
controller records both the commanded 20-DoF joint targets and the resulting
joint states. Each demonstration therefore carries the actions actually sent to
the hand, rather than actions inferred from measured states,
satisfying the \textbf{Action Data} requirement. The rigid connections keep the controller-to-hand and camera-to-hand transforms fixed, so the tracked motion and visual observations
remain expressed relative to the same hand frame.

\noindentparagraph{Ergonomic Design.}
The complete hand-held assembly weighs \(0.955\)\,kg, of which the non-hand
hardware contributes \(0.220\)\,kg. To support this load while leaving the thumb
and index finger free to operate the controls, we extend the Quest 2 battery
cover into a collection handle that connects the controller to the transferable
module. A strap across the back of the operator's hand carries part of the load,
reducing the sustained grip force.

Our initial design placed the controller behind the robot hand, offsetting the
operator's grip from the assembly's center of mass and creating a long moment
arm, as shown in \cref{fig:primitive-mapping}(b). The final design positions the
controller above the hand so that the grip lies near the center of mass. This
shortens the moment arm and reduces the wrist torque that the operator must
counteract.

\subsection{Dexterous Control via Parameterized Skills}
\label{sec:skills}

Control via skills is an effective way to operate high-dimensional robotic
systems. For example, a humanoid can be commanded through a desired velocity
rather than by specifying the motion of every joint. We apply the same principle
to dexterous hands by separating the operator's intent from the finger
motions required to realize it. To this end, we design a compact set of skills that we find effective across a wide range of dexterous tasks. The skill interface is \textit{parameterized} instead of discrete, allowing more wholistic control. For instance, the user controls the strength of the grasp by using the joystick to incrementally close or open the grasp, instead of snapping to a single fixed grasp.

Together, the skills keep all 20 joints of the hand available during collection without
asking the operator to command each one, satisfying the \textbf{DoF} 
requirement that separates a dexterous interface from a simpler gripper.
We implement this interface using a Meta Quest 2 controller, which provides two
buttons, two continuous triggers, and a two-axis joystick. As illustrated in
\cref{fig:primitive-mapping}(c) and detailed below, these inputs allow the operator to select and modulate the skills.

\begin{itemize}[\setlength{\labelindent}{0.3em}%
                \setlength{\labelsep}{0.35em}]
\item \textbf{Thumb Control.} The two joystick axes control thumb flexion and
extension along one axis and abduction and adduction along the other, following
HATO~\cite{hato}. This allows the operator to position the thumb independently
for different grasp geometries.
\item \textbf{Power Grasp.} One joystick axis controls the coordinated closure
of the four non-thumb fingers. Moving the joystick incrementally loosens or
tightens the grasp, while releasing it holds the current target. Unlike
continuous trigger control~\cite{hato}, this persistent command does not require
the operator to maintain pressure throughout a grasp. It is primarily used to
wrap the hand around larger objects.
\item \textbf{Pinch Grasp.} The same incremental command controls only the index
finger, which closes against the independently positioned thumb. This skill
supports small objects and grasps that require precise fingertip contact.
\item \textbf{Two-Finger Grasp.} The index and middle fingers open or close
together while the remaining fingers retain their targets. This skill
supports multi-stage behaviors such as multi-object grasping and two-finger
grasp-and-pull motions, without requiring the operator to coordinate the two
fingers individually.
\item \textbf{In-Hand Rotation.} A button invokes a learned closed-loop policy that
generates the repeated finger gaiting required to rotate a grasped tool in hand.
The policy controls the fingers while the operator continues to control the
global motion of the hand.
\end{itemize}

\noindentparagraph{Learning the In-Hand Rotation Skill.}
We learn the in-hand rotation skill in simulation.
This relieves the operator of difficult low-level, in-hand manipulation, 
especially without finger-level haptic feedback. Similar to~\cite{dexscrew},
we train a closed-loop screwing policy in simulation with PPO~\cite{ppo}
in Isaac Lab~\cite{isaaclab}. 

We model the task as rotating a cylindrical handle on a revolute joint. In the real-world, the operator can make wrist adjustments, but we simplify the simulation environment so hand base is fixed, with the policy commanding fingers only. The policy learns delta actions over 12 actuated joints, with the ring and pinky finger fixed. The policy observes a three-step history, including measured joint positions, commanded targets, and the previous action. We learn a privileged critic which has access to the handle's rotation angle. 

We follow rewards and domain randomization similar to ~\cite{dexscrew}, which includes a handle rotation, fingertip proximity, bounded action smoothness, initial grasp proximity, joint torque, and mechanical work rewards. We randomize object mass, center of mass, friction, restitution, finger PD gains, initial grasp, and random disturbance forces. We train for $1.18 \times 10^{8}$ environment steps across $4096$ environments. During collection, the actor, a three-layer MLP, runs at $20$\,Hz on the hand's joint encoders while the operator retains control of the wrist.

\subsection{Policy Learning}
\label{sec:policy}

\noindentparagraph{Raw-Action Supervision.}
The skills are used only to assist the operator during data
collection. The learned policy directly predicts wrist poses and 20-DoF hand joint targets. At each
time step, we record the raw 20-DoF target produced by the active skill and
sent to the hand controller, together with the resulting measured joint state.
The policy receives no skill identity, skill parameter, or task-phase
label. Instead, it is trained to predict the commanded joint targets directly.
Thus, the skills structure how demonstrations are collected without
constraining the representation learned by the policy.
Switching the action label from the commanded target to the measured joint state is the single
change that defines our state-as-action baseline in \cref{sec:experiments}; every other part of the
recipe below is shared.

\noindentparagraph{Observation and Action Spaces.}
We train a diffusion policy~\cite{dp} with a convolutional U-Net denoiser and a
fine-tuned DINOv2 visual encoder. Images are augmented with a random crop and resize, color jitter, random grayscale, and Gaussian blur. The vision encoder is optimized with a reduced learning
rate, one-tenth that of the remaining network. The policy observes the two most
recent time steps, each consisting of a $224\times224$ RGB image, the
end-effector position represented relative to the current chunk start~\cite{umi}, and
the measured 20-DoF hand state. 

The policy predicts a 16-step action chunk containing
end-effector poses relative to the start of the chunk and absolute 20-DoF hand
targets. Each action step is $29$-dimensional: $9$ for the wrist ($3$ translation and a 6D
rotation) and $20$ for the hand. For bimanual tasks, we concatenate the observations and actions of the two hands without otherwise changing the architecture. We train 50 DDIM steps and sample with 10 steps. Following HATO~\cite{hato}, policy inference
runs asynchronously to maintain smooth control.

\noindentparagraph{Deployment Calibration.}
The collected end-effector actions are expressed in the Quest controller frame,
whereas deployment actions must be expressed in the robot end-effector frame.
We estimate the fixed transform between these frames from paired controller and
end-effector poses collected with the device mounted on the robot. Concretely, we bolt the module to the robot flange and record $N$ paired samples of the flange
pose $R_i$, and the controller pose $Q_i$, from
the headset. Each sample satisfies $Q_i = X R_i M$, where $X$ is the unknown robot base to headset
transform and $M$ is the controller to flange transform, which is the mount transform we want. Any two samples eliminate $X$,
\begin{equation}
    R_i^{-1} R_j = M \, (Q_i^{-1} Q_j) \, M^{-1},
    \label{eq:mount}
\end{equation}
so $N$ poses give up to $N(N-1)/2$ pairwise constraints on exactly the conjugation that deployment applies,
$A^{\text{ee}} = M A^{\text{quest}} M^{-1}$. We fit the full 6-DoF $M$ by least squares on
\cref{eq:mount}, independently for each of the two arms. We apply
this transform to the predicted end-effector motions at deployment; the 20-DoF
hand targets require no conversion. 

%% file: sections/4_experiments.tex
\section{EXPERIMENTS}
\label{sec:experiments}

\begin{figure}[t]
    \vspace{6pt}
    \centering
    \includegraphics[width=\linewidth]{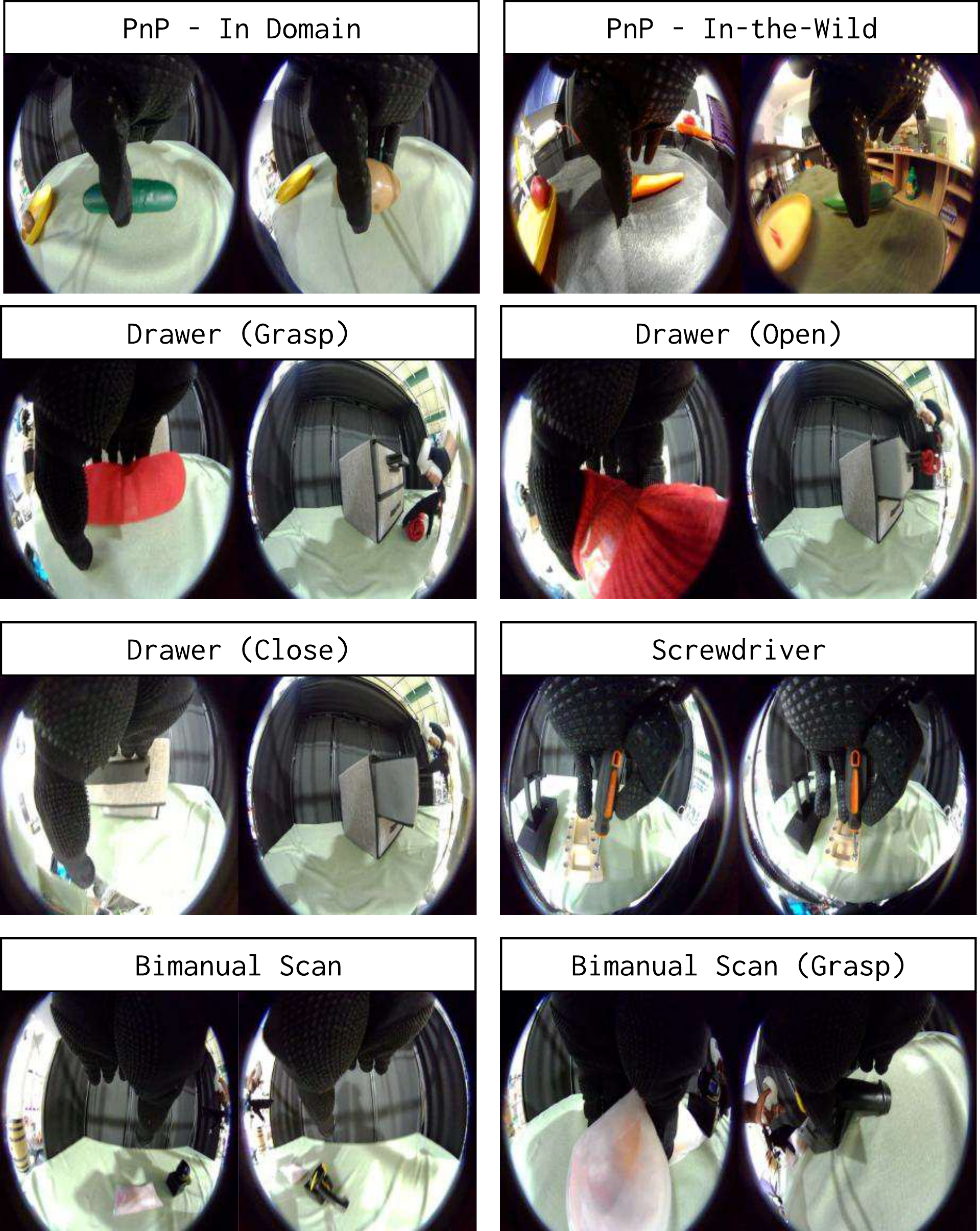}
    \caption{\textbf{Data Visualization.} We show fisheye camera views recorded during data collection. For our pick-and-place task (PnP), we collect in-domain and in-the-wild data. For drawer stowing, we add a third-person camera because the grasped object occludes much of the scene. We show both camera views at three stages of the trajectory. For bimanual scanning and drawer stowing, the operator is occasionally visible, but we do not post-process the data to remove the operator.
    }
    \label{fig:fisheye}
    \vspace{-10pt}
\end{figure}

\begin{figure*}[t]
    \vspace{6pt}
    \centering
    \includegraphics[width=\linewidth]{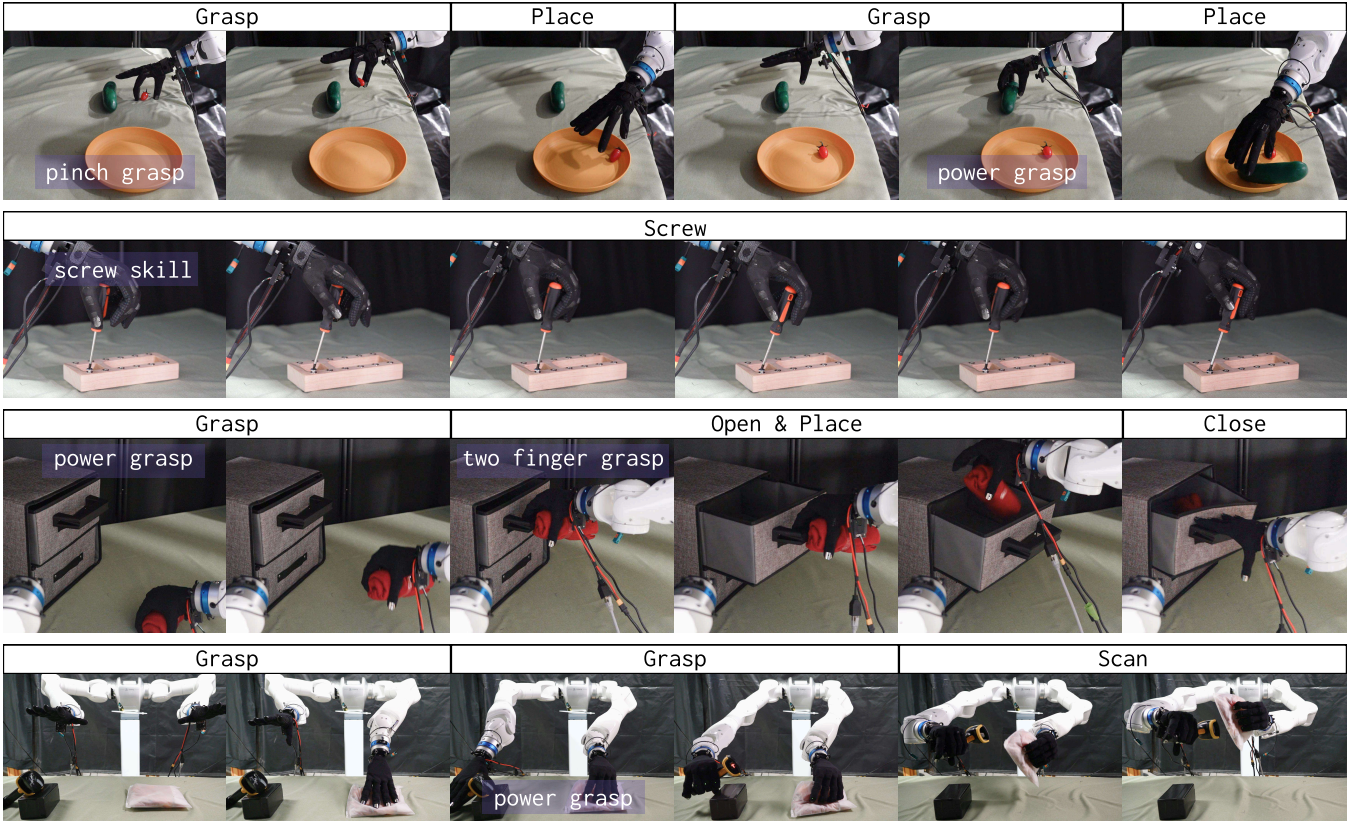}
    \caption{\textbf{Task Rollouts.} We show successful rollouts annotated with the phases of each task. We use these phases to score partial success. We also annotate grasp types that emerge in the rollouts, such as a pinch grasp for the cherry tomato and a power grasp for the cucumber. These annotations describe observed behavior. The policy predicts raw joint targets without skill labels.
    }
    \label{fig:rollouts}
    \vspace{-3pt}
\end{figure*}

\begin{figure*}[t]
    \centering
    \includegraphics[width=\linewidth]{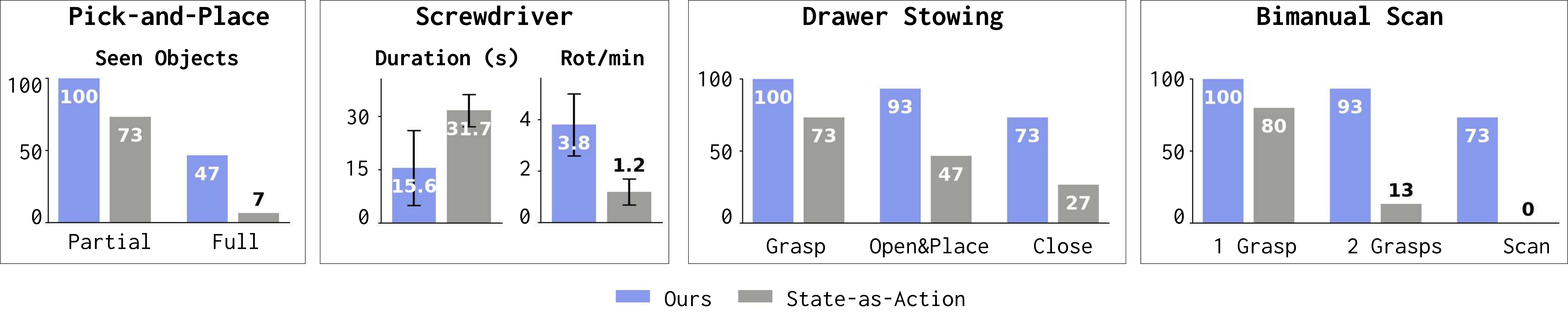}
    \vspace{-2em}
    \caption{\textbf{Policy Learning Results.} We show imitation learning performance across four tasks. Policies trained on recorded hand commands achieve higher success rates and faster screwdriver rotation than the state-as-action baseline. Partial success for pick-and-place denotes grasping and placing one object. For drawer stowing and bimanual scanning, we report success rates for the task phases shown in~\cref{fig:rollouts}. A phase counts as successful only if all preceding phases also succeed. Each policy is evaluated for 15 episodes per task.
    }
    \label{fig:results}
    \vspace{-16pt}
\end{figure*}

We evaluate \method through three research questions:
\begin{itemize}[\setlength{\labelindent}{0em}%
                \setlength{\labelsep}{0.35em}]
\item \textbf{Q1:} Do the hand commands recorded by \method enable policies that
succeed on diverse dexterous tasks?
\item \textbf{Q2:} Does co-training with demonstrations collected in unstructured
environments improve robustness to unseen objects and cluttered scenes?
\item \textbf{Q3:} How does the collection throughput of \method compare with
teleoperation and human-hand demos?
\end{itemize}

\subsection{Experimental Setup}

\noindentparagraph{Robot Platform.}
We deploy policies on Tianji Marvin, a bimanual robot with two 7-DoF arms. The
hand--camera modules used for collection (\cref{sec:hardware}) mount directly to
its wrists, so the hand embodiment and camera-to-hand extrinsics are identical
across collection and deployment. Hand targets therefore need no retargeting, and
the only correction at deployment is the fixed controller-to-end-effector
transform (\cref{sec:policy}).

We showcase the modularity of our system by using different hands: the Wuji 1 hand for Pick-and-Place and Screwing, and the Wuji 2 hand for Drawer Stowing and Bimanual Barcode Scanning.

\noindentparagraph{Tasks.}
We evaluate four tasks that exercise complementary forms of dexterity.
\cref{fig:rollouts} shows a successful rollout of each, annotated with the phases
we use to score partial success and with the grasp types that emerge:
\begin{itemize}[\setlength{\labelindent}{0em}%
                \setlength{\labelsep}{0.35em}]
\item \textbf{Vegetable Pick-and-Place.} The robot sequentially picks and places
two vegetables into a plate. Variations in object size and shape require the
policy to select between pinch and power grasps without explicit grasp labels.
\item \textbf{Screwing.} The robot maintains a stable grasp while rotating a
screwdriver in hand, testing sustained contact and repeated finger gaiting.
\item \textbf{Drawer Stowing.} Without releasing a grasped object, the robot
opens a drawer with the same hand, places the object inside, and closes the
drawer. This task couples object retention with articulated-object manipulation.
\item \textbf{Bimanual Barcode Scanning.} One hand grasps and presents an object
while the other positions and operates a barcode scanner. Success requires
coordinated bimanual positioning of the object and scanner.
\end{itemize}

\noindentparagraph{Observations.}
Policies observe the fisheye view from the hand--camera module, shown for every
task in \cref{fig:fisheye}. For drawer stowing we add a third-person camera,
because the grasped object occludes much of the scene while the drawer is pulled
open. The operator is occasionally visible in the recorded frames, and we apply no
post-processing.


\noindentparagraph{Datasets.}
We collect 1,605 laboratory demonstrations across the four tasks: 700 for
vegetable pick-and-place, 200 for screwdriver manipulation, 380 for drawer
stowing, and 325 for bimanual barcode scanning. We additionally collect 1,000
pick-and-place demonstrations in unstructured apartment and office environments.
The co-trained policies use both the laboratory and in-the-wild pick-and-place
datasets.

\noindentparagraph{Training Variants and Baselines.}
To answer Q1, we compare policies trained on the commanded hand targets recorded
by \method against a state-as-action baseline that instead uses measured hand
states as action labels. To answer Q2, we compare the laboratory-only
pick-and-place policy with two co-trained variants. To answer Q3, we compare the
number of trajectories collected per 30 minutes using \method, teleoperation,
and human-hand demonstrations.

\noindentparagraph{Evaluation Protocol.}
We evaluate each policy for 15 episodes per task or evaluation condition.
For vegetable pick-and-place, we test seen objects, unseen objects, and scenes
with distractors; partial success denotes placing one of the two objects, while
full success denotes placing both. For drawer stowing and bimanual scanning, we
report completion of the task phases visualized in \cref{fig:rollouts}. For
screwdriver manipulation, we report the average hold duration in seconds and rotations per
minute.

\begin{figure}[t]
    \vspace{6pt}
    \centering
    \includegraphics[width=0.8\linewidth]{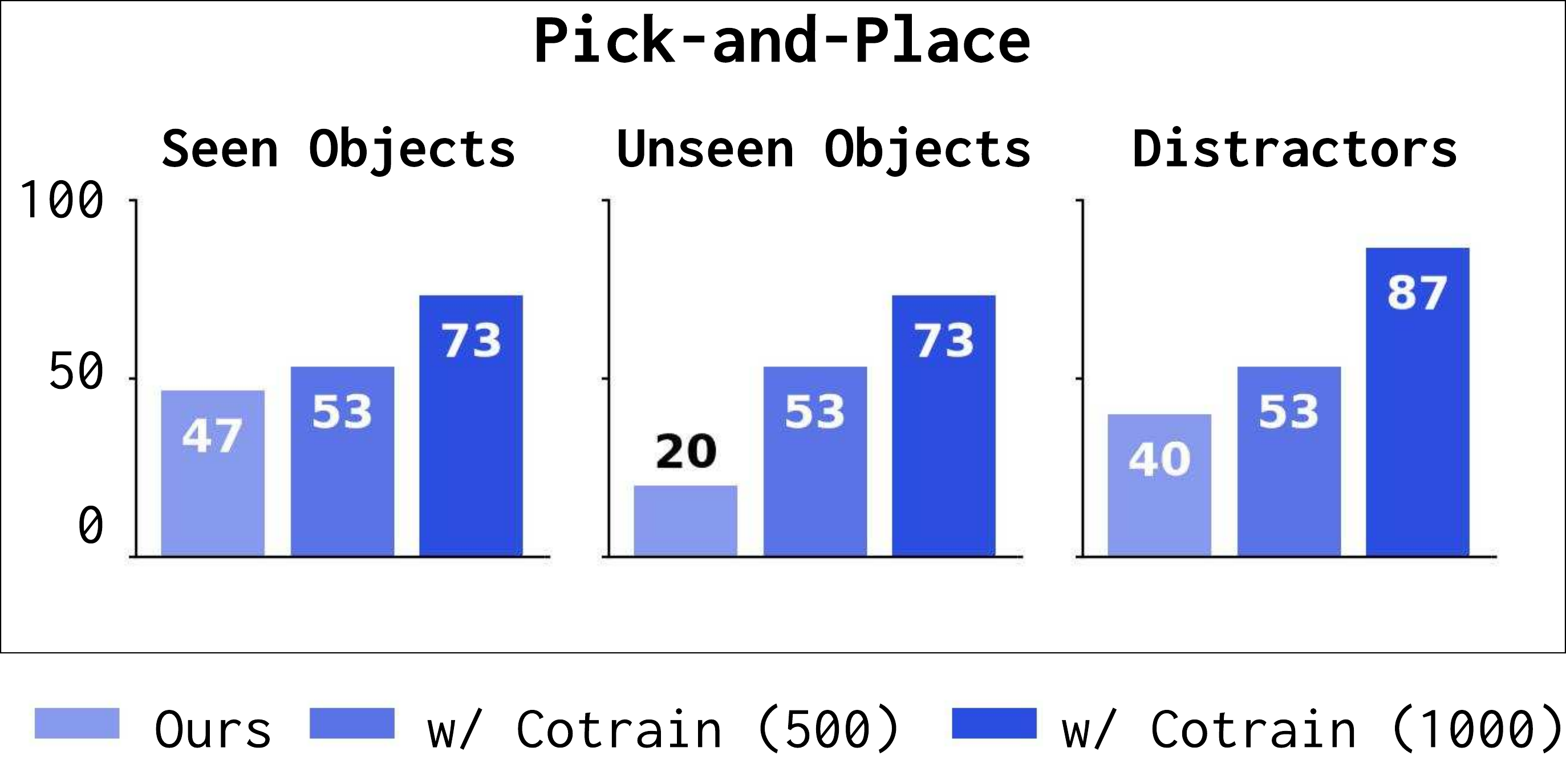}
    \caption{\textbf{In-the-Wild Co-Training.} One benefit of \method's design is its portability. For our pick-and-place task, we collect 1,000 in-the-wild demonstrations (visualized in~\cref{fig:fisheye}) and co-train with the 700 laboratory demonstrations. We find that increasing the number of cotrained trajectories increases the success rate of the policy in every condition. In the distractor setting, full success rises from 40\% to 87\%.
    }
    \label{fig:results_wild}
    \vspace{-10pt}
\end{figure}

\subsection{Policy Learning Across Dexterous Tasks (Q1)}

\cref{fig:results} reports imitation learning results across the four tasks.
Policies trained on \method demonstrations acquire all of the target behaviors:
they select between pinch and power grasps without grasp labels, hold an object
while pulling open a drawer with the same hand, gait the fingers to rotate a
screwdriver in hand, and coordinate two arms to present an object to a scanner.

\begin{figure}[t]
    \vspace{6pt}
    \centering
    \includegraphics[width=\linewidth]{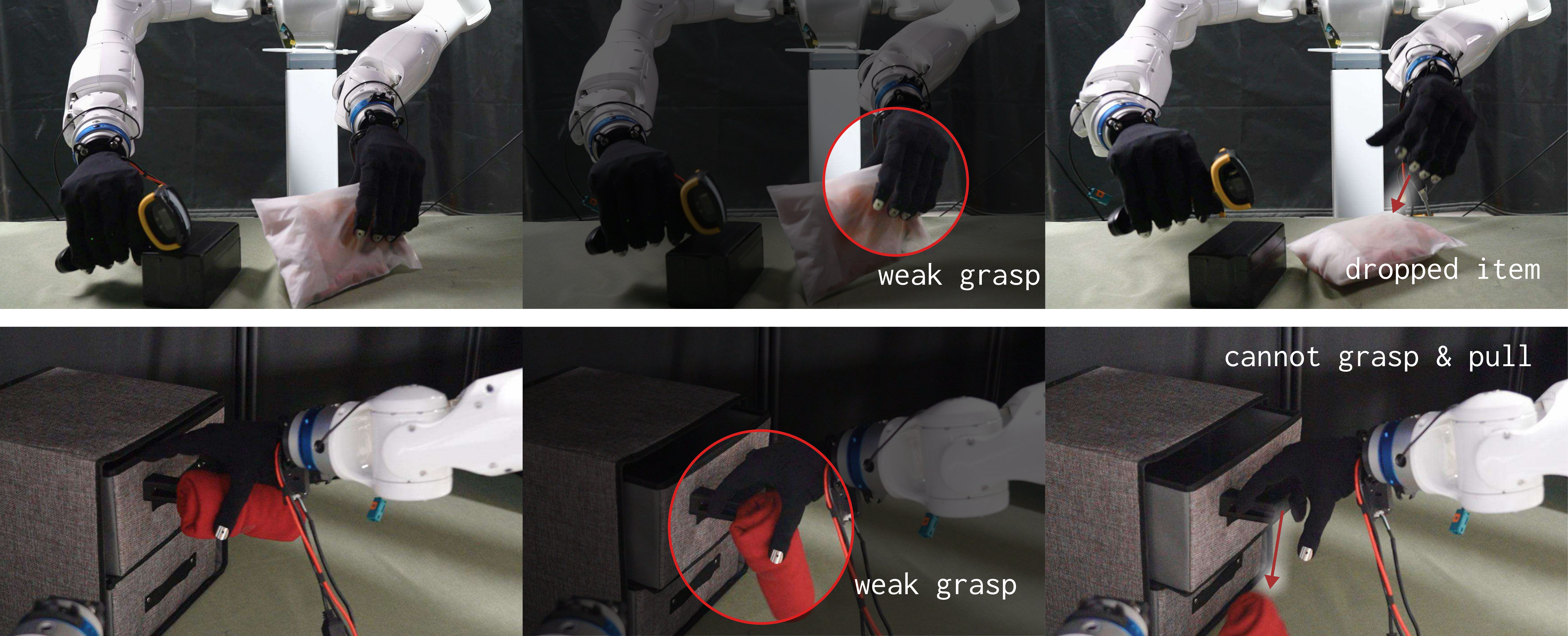}
    \caption{\textbf{State-as-Action Failure.} Policies
    trained with measured joint states as action labels produce weak grasps and fail to hold onto objects. \textit{Top:} the baseline drops the package when
    presenting it to the scanner. \textit{Bottom:} the baseline cannot retain the laundry
    once it begins to pull the drawer open.
    }
    \label{fig:state_fail}
    \vspace{-16pt}
\end{figure}

The state-as-action baseline, which uses measured joint states as action labels,
is worse on every task. For instance, we find that the state-as-action baseline is unable to sufficiently grasp the object. For the scanning task, grasping the package is nearly impossible for the baseline, whereas with robot commands, we exert the necessary force to hold the item tightly. For the drawer task, the laundry often falls during the drawer opening segment. We visualize this in \cref{fig:state_fail}.
Because a stage cannot succeed once an earlier stage has failed, the deficit
compounds rather than appearing as a constant offset: on bimanual scanning the
baseline grasps the first object in 80\% of episodes, holds both in 13\%, and
never completes a scan. The screwdriver task exposes the same weakness through a
different metric. While the baseline holds the tool more than twice as long as our
policy (31.7\,s vs.\ 15.6\,s), it rotates it at a third of the rate and is much less effective. These results highlight the importance of \textbf{Action Data}, since access to robot commands for the deployment hand is crucial for a higher success rate. This is a major limitation for data collection interfaces based on exoskeletons or hardware without action data.

\subsection{Co-Training with In-the-Wild Data (Q2)}
We collect a large in-the-wild pick-and-place dataset across several sites,
including an apartment room and a range of office tables. \method
requires neither a robot arm nor external tracking infrastructure, thus enabling this portable data collection. We
use the same training objects,
with the difference being diversity in scene, background, and lighting rather than in
object identity, as \cref{fig:fisheye} illustrates.

\cref{fig:results_wild} shows that full success improves in every condition and
grows with the amount of in-the-wild data. With 1,000 in-the-wild demonstrations, full success rises from 40\%
to 87\% under distractors and from 20\% to 73\% on unseen objects, so the
benefit transfers to new objects. It also improves on seen objects, from 47\%
to 73\%. \cref{app:wild_scaling} reports a lower
data regime.

\begin{table}[t]
    \vspace{6pt}
    \centering
    \small
    \setlength{\tabcolsep}{5pt}
    \resizebox{\columnwidth}{!}{%
    \begin{tabular}{lccc|c}
    \toprule
& \method & Skill-Teleop & Glove-Teleop & Human hand \\
    \midrule
    Traj. / 30\,min  & 122        & 68         & 86         & 275        \\
    Portable         & \Checkmark & \XSolidBrush & \XSolidBrush & \Checkmark \\
    Same hand emb.   & \Checkmark & \Checkmark & \Checkmark & \XSolidBrush \\
    Action data      & \Checkmark & \Checkmark & \Checkmark & \XSolidBrush \\
    \bottomrule
\end{tabular}}
    \vspace{0.5\baselineskip}
    \caption{\textbf{Data Collection Throughput.} Trajectories collected during
    30 minutes of continuous operation on the unimanual pick-and-place task. For Glove Teleoperation, we use the Wuji Glove for hand tracking and the VR controller for end-effector control. For Skill Teleoperation, we use the same skill interface as \method to command the hand and VR tracking to control the end effector. \method is the only method compared here
    that combines portable collection with the deployed hand's embodiment and its
    commanded actions.
    }
    \label{tab:throughput}
    \vspace{-20pt}
\end{table}

\subsection{Data Collection Throughput (Q3)}
We measure throughput as the number of trajectories collected during 30 minutes of
continuous operation on the unimanual pick-and-place task, comparing \method
against bare-hand human demonstration and teleoperation. For Glove Teleoperation, the operator uses the Wuji Glove for hand tracking and the VR controller for the end-effector. For
Skill Teleoperation, we use the VR controller for end-effector control and the same hand
skills as \method. The operator is expert-level for each style of data collection.

\cref{tab:throughput} shows that \method collects roughly $1.4\times$ as many trajectories as Glove Teleoperation and $1.8\times$ as many
trajectories as Skill Teleoperation. While Skill Teleoperation is a simpler interface, reducing the operator's cognitive burden, Glove Teleoperation still remains faster because the hand tracking is more responsive and flexible. Generally, both are much slower than \method. When teleoperating, the operator is out of the scene, removing the direct visual and haptic cues. This makes approaching an object and performing precise tasks, such as pinch grasping a small object, much slower and more difficult. Human hand data collection remains about $2.3\times$ faster than
\method, but we regard it as an upper bound on speed rather than a competing
source of training data, as human hand data provides neither the deployed hand's embodiment nor its commanded actions.

%% file: sections/5_conclusion.tex
\section{CONCLUSIONS }

We presented \method, a portable interface for collecting demonstrations with
high-DoF dexterous hands. The system uses the same hand hardware for collection and
deployment, preserving its contact geometry while recording its commanded
targets and resulting joint states. A compact set of skills makes the hand
practical to operate by separating task-level intent from detailed finger
coordination. These skills assist only the demonstrator; the learned policy
directly predicts the raw hand commands without skill or task-phase labels.
In our experiments, \method collects 122 trajectories per 30 minutes, roughly
$1.4\times$ the rate of teleoperation. Policies trained on the resulting
demonstrations succeed on all four tasks and consistently outperform a
state-as-action baseline, which never completes a bimanual barcode scan.
With 700 laboratory demonstrations, co-training with in-the-wild data further
improves pick-and-place success in every evaluation condition.
Together, these design choices provide a practical and scalable solution for
collecting portable, executable demonstrations for dexterous robot learning.

\noindentparagraph{\textbf{Limitations.}} Despite its ergonomic design, the weight of the device limits
uninterrupted use and rotation-heavy motions. As the weight of the hand is the primary bottleneck, we expect that future hardware developments for lighter and smaller hands can address this. Our skills also cover
only part of the space of dexterous behaviors, and future work may explore extending them with additional
skills learned in simulation, such as other in-hand manipulation skills, learned grasps, and more.

\section*{ACKNOWLEDGMENT}
The authors would like to thank Keyshawn King and Volodymyr Behma for data collection and support, and Andrew Wright for initial hardware designs. We also thank Tingfan Wu, Anas Houssaini, Jensen Gao, and the FAR team for feedback and support. 

%% file: sections/6_appendix.tex
%
\appendix
\crefalias{subsection}{appendix}
\suppressfloats[t]

\subsection{Scaling In-the-Wild Data}
\label{app:wild_scaling}

\begin{table}[t]
    \vspace{6pt}
    \centering
    \small
    \setlength{\tabcolsep}{4pt}
    \begin{tabular}{ccccc}
        \toprule
        & & \multicolumn{3}{c}{Success (\%)} \\
        \cmidrule(lr){3-5}
        \# Demos & \# In-the-Wild & Seen & Unseen & Distractors \\
        \midrule
        500 & 0     & 0  & 20 & 0  \\
        500 & 500   & 33 & 13 & 53 \\
        500 & 1,000 & 47 & 7  & 40 \\
        \midrule
        700 & 0     & 47 & 20 & 40 \\
        700 & 500   & 53 & 53 & 53 \\
        700 & 1,000 & 73 & 73 & 87 \\
        \bottomrule
    \end{tabular}
    \vspace{0.5\baselineskip}
    \caption{\textbf{Scaling In-the-Wild Data.} Pick-and-place success rates (\%)
    of policies trained on 500 or 700 in-scene laboratory demonstrations (\# Demos) and 0,
    500, or 1,000 in-the-wild demonstrations (\# In-the-Wild). We evaluate on seen
    objects, unseen objects, and scenes containing distractors. In-the-wild data
    improves full success on seen objects and with distractors in both data regimes, but improves success on unseen objects only with 700 laboratory demonstrations.
    }
    \label{tab:wild_scaling}
    \vspace{-6pt}
\end{table}

\cref{fig:results_wild} shows the benefit of scaling in-the-wild data with 700 laboratory demonstrations. We also evaluate pick-and-place policies trained on 500 laboratory demonstrations with 0, 500, or 1,000 in-the-wild demonstrations.

From \cref{tab:wild_scaling}, co-training with in-the-wild data improves full success on seen objects and with distractors for both the 500 in-scene and 700 in-scene demonstration settings. However, for 500 demos, in-the-wild data does not improve performance in the unseen object evaluation setting. Because the in-the-wild demonstrations use
the same objects as the laboratory data, they diversify scenes rather than
object geometry, and we hypothesize that transferring this robustness to new
objects requires a sufficiently large laboratory dataset.

\noindentparagraph{Training Details.} For the policies, we cotrain with 50\% of the batch drawn from in-the-wild data. We train every policy for the same number of epochs over its
laboratory demonstrations ($\approx$250 epochs).

\subsection{Hardware and Data Collection}
\label{app:collection}

\noindentparagraph{Hardware.} The hand--camera module pairs a Wuji hand with a
USB board camera with a 200$^\circ$ fisheye lens.
On the Wuji~1, hand targets pass through a
5\,Hz low-pass filter and are interpolated by the hand firmware at 1\,kHz, with
a per-joint torque limit of 0.5\,N$\cdot$m.

\noindentparagraph{Recording and Synchronization.} A single recording loop runs
at 30\,Hz. The camera, hand, and Quest controller are each
read in a background thread. Because the camera and the hand use same 30\,Hz loop during collection and deployment, we expect their
latencies to roughly match across the two settings.  We therefore do not measure stream latencies or temporally align
the streams in post-processing, as UMI~\cite{umi} does. At deployment, the
asynchronous inference in \cref{app:deployment} discards actions whose time
has already passed.

\subsection{In-Hand Rotation Skill}
\label{app:rotation}

\begin{table}[t]
    \vspace{6pt}
    \centering
    \small
    \resizebox{\columnwidth}{!}{%
    \begin{tabular}{ll}
        \toprule
        Hyperparameter & Value \\
        \midrule
        RL library                  & RSL-RL~\cite{rslrl} \\
        Parallel environments       & $4096$ \\
        Rollout length              & $24$ steps \\
        Iterations                  & $1200$ ($1.18\times10^{8}$ steps) \\
        Learning rate               & $5\times10^{-3}$, adaptive (KL target $0.01$) \\
        Discount $\gamma$ / GAE $\lambda$ & $0.99$ / $0.95$ \\
        Clip ratio                  & $0.2$ \\
        Value loss coefficient      & $1.0$ (clipped) \\
        Entropy coefficient         & $0$ \\
        Epochs / minibatches        & $5$ / $4$ \\
        Max gradient norm           & $1.0$ \\
        Initial action std.         & $1.0$ \\
        Actor and critic MLPs       & $[512, 256, 128]$, ELU \\
        Observation normalization   & Running mean and std. \\
        Control rate / episode      & $20$\,Hz / $40$\,s \\
        \bottomrule
    \end{tabular}}
    \vspace{0.5\baselineskip}
    \caption{\textbf{PPO Hyperparameters.} Settings for training the in-hand
    rotation skill in Isaac Lab~\cite{isaaclab}. The simulation runs at 200\,Hz
    with 10 physics steps per policy step.
    }
    \label{tab:ppo}
    \vspace{-6pt}
\end{table}

\begin{table}[t]
    \vspace{6pt}
    \centering
    \small
    \setlength{\tabcolsep}{4pt}
    \begin{tabular}{lcl}
        \toprule
        Term & Weight & Definition \\
        \midrule
        Handle rotation     & $6.0$   & $\mathrm{clip}(\dot\theta, -4, 4)$ \\
        Fingertip proximity & $2.0$   & $\mathrm{clip}(1 - \bar d / 0.035, 0, 1)$ \\
        Action smoothness   & $1.0$   & $\exp\!\big(-\overline{(a_t - a_{t-1})^2} / 0.5^2\big)$ \\
        Pose deviation      & $-6.0$  & $\lVert q - q_0 \rVert^2$ \\
        Over-speed          & $-0.3$  & $\max(\dot\theta - 10, 0)$ \\
        Joint torque        & $-0.1$  & $\lVert \tau \rVert^2$ \\
        Mechanical work     & $-0.01$ & $\big(\sum_j \lvert \tau_j \rvert \lvert \dot q_j \rvert\big)^2$ \\
        \bottomrule
    \end{tabular}
    \vspace{0.5\baselineskip}
    \caption{\textbf{Rotation Skill Rewards.} $\dot\theta$ is the handle's
    angular velocity in rad/s, $\bar d$ is the mean horizontal distance in meters
    from the thumb, index, and middle fingertips to the handle axis, $q$ and $q_0$
    are the current and initial joint position vectors, and $\tau$ is the joint torque vector.
    $a_t$ is the policy action at step $t$, $\dot q_j$ is the velocity of joint $j$,
    and the overline in the action smoothness term denotes the mean over joints.
    }
    \label{tab:rewards}
    \vspace{-6pt}
\end{table}

\noindentparagraph{Environment.} We model the screwdriver as a 12-sided prism that rotates about its
axis on a revolute joint. The hand base
is fixed in simulation. An episode terminates early
if the thumb, index, or middle fingertip moves more than 12\,cm from the handle.
\cref{tab:ppo,tab:rewards} list the training hyperparameters and reward
terms, and \cref{sec:skills} lists the randomized parameters.

\noindentparagraph{Observations and Actions.} The actor observes a 147-D vector:
the 20 measured joint positions and 20 joint targets over the last three steps,
the previous action, and the fixed pose of the hand base. Following the
asymmetric actor-critic approach~\cite{asymac}, the critic observes a 127-D
privileged state without history: joint positions, targets, and velocities,
the hand base pose, fingertip positions, the handle's position, rotation angle,
and angular velocity, the applied joint torques, and the previous action. The
policy outputs a 20-D action $a \in [-1, 1]^{20}$, and each joint target moves
by $0.04\,a$ rad per step within the joint limits, which is a scaling factor inspired by~\cite{dexscrew}. The ring and pinky fingers are fixed, and the policy
controls the remaining 12 joints.

\noindentparagraph{Real-World Execution.} During collection, the actor runs at
20\,Hz on joint positions measured by the hand's encoders and holds its targets
between policy steps, while the operator controls the wrist through the
collection handle. 

\subsection{Policy Learning}
\label{app:policy}

\begin{table}[t]
    \vspace{6pt}
    \centering
    \small
    \setlength{\tabcolsep}{4pt}
    \resizebox{\columnwidth}{!}{%
    \begin{tabular}{ll}
        \toprule
        Hyperparameter & Value \\
        \midrule
        Vision encoder            & DINOv2 ViT-S/14~\cite{dinov2}, fine-tuned \\
        Denoiser                  & 1D conditional U-Net~\cite{dp}, FiLM~\cite{film} \\
        U-Net channels per level  & \\
        \quad Pick-and-place in-the-wild           & $(512, 1024, 2048)$  \\
        \quad Pick-and-place baseline  & $(256, 512, 1024)$ \\
        \quad Screwdriver              & $(512, 1024, 2048)$ \\
        \quad Drawer stowing           & $(512, 1024, 2048)$ \\
        \quad Bimanual scanning        & $(512, 1024, 2048)$ \\
        \midrule
        Noise scheduler           & DDIM~\cite{ddim}, squared cosine \\
        Prediction target         & Noise ($\epsilon$) \\
        Train / inference steps   & $50$ / $10$ \\
        \midrule
        Optimizer                 & AdamW, $\beta = (0.9, 0.999)$ \\
        Weight decay              & $0.01$ \\
        Learning rate             & $1\times10^{-4}$ (vision encoder $1\times10^{-5}$) \\
        LR schedule               & Cosine, $500$ warmup steps \\
        Batch size                & $64$ \\
        Gradient steps            & $5\times10^{5}$ \\
        EMA decay                 & $\min(1 - (1 + t)^{-3/4},\ 0.9999)$ \\
        \midrule
        Random crop               & $212$ px, resized to $224$ \\
        Color jitter              & $(0.3, 0.4, 0.5, 0.08)$ \\
        Random grayscale          & $p = 0.1$ \\
        Gaussian blur             & $k = 5$, $\sigma \in [0.1, 1.5]$, $p = 0.2$ \\
        \bottomrule
    \end{tabular}}
    \vspace{0.5\baselineskip}
    \caption{\textbf{Diffusion Policy Hyperparameters.} Settings are shared across
    tasks except for the task-specific U-Net channel sizes listed above. Color jitter lists the
    brightness, contrast, saturation, and hue ranges. At inference, images are
    center-cropped to 212\,px and resized to 224\,px.
    }
    \label{tab:dp}
    \vspace{-6pt}
\end{table}

\noindentparagraph{Architecture.} \cref{tab:dp} lists the hyperparameters of
the diffusion policy in \cref{sec:policy}. The end-effector pose is
represented relative to the current end-effector frame, with a 3-D translation
and a 6-D rotation. Each predicted wrist action is a relative pose with respect
to the start of the chunk, and each hand action is an absolute 20-D target.
Bimanual policies concatenate the observations and actions of both hands and
additionally observe the relative pose between them. All inputs and targets are
normalized using their 2nd and 98th percentiles.

\subsection{Deployment}
\label{app:deployment}

\noindentparagraph{Mount Calibration.} To estimate the controller-to-flange
transform $M$ in \cref{eq:mount}, we bolt the module to each robot flange and
jog the arm through about 30 poses that tilt the flange axis by at least
30$^\circ$, pausing 1.5\,s at each pose before recording the paired flange and
controller poses. We initialize $M$ with a closed-form hand--eye solution and
refine it by nonlinear least squares.

\noindentparagraph{Control.} The policy runs at 30\,Hz, matching the collection
rate. Following HATO~\cite{hato}, inference runs asynchronously: each new
16-step chunk is aligned to the control tick at which its observation was
taken, the first 8 steps are kept, steps whose time has passed are dropped, and
the hand and wrist commands at each tick average all chunks that cover it. We
sample each chunk with 10 DDIM steps. Predicted wrist motions are converted to
flange targets with the mount transform and solved for the 7-DoF arm with
Levenberg--Marquardt inverse kinematics in PyRoki~\cite{pyroki}.